\documentclass[10pt,twocolumn,letterpaper]{article}

\usepackage{cvpr}
\usepackage{times}
\usepackage{epsfig}
\usepackage{graphicx}
\usepackage{amsmath}
\usepackage{amssymb}
\usepackage{subcaption}
\usepackage[group-separator={,}]{siunitx}
\usepackage{bm}
\usepackage{booktabs}       %
\graphicspath{ {./figures/} }

\usepackage[pagebackref=true,breaklinks=true,letterpaper=true,colorlinks,bookmarks=false]{hyperref}

\cvprfinalcopy %

\def\cvprPaperID{12} %

\ifcvprfinal\fi
\begin{document}

\title{ePillID Dataset: A Low-Shot Fine-Grained Benchmark for Pill Identification}

\author{
	Naoto Usuyama, Natalia Larios Delgado, Amanda K. Hall  \\
	Microsoft Healthcare \\
	{\tt\small\{naotous,nlarios,amhal\}@microsoft.com}
	\and
	Jessica Lundin \\
	Work done at Microsoft\\
	{\tt\small lundinmachine@outlook.com }
}

\maketitle
\thispagestyle{empty}

\begin{abstract}
Identifying prescription medications is a frequent task for patients and medical professionals; however, this is an error-prone task as many pills have similar appearances (e.g. white round pills), which increases the risk of medication errors. In this paper, we introduce ePillID, the largest public benchmark on pill image recognition, composed of 13k images representing 9804 appearance classes (two sides for 4902 pill types). For most of the appearance classes, there exists only one reference image, making it a challenging low-shot recognition setting.
We present our experimental setup and evaluation results of various baseline models on the benchmark. The best baseline using a multi-head metric-learning approach with bilinear features performed remarkably well; however, our error analysis suggests that they still fail to distinguish particularly confusing classes. \footnote{The code and data are available at \url{https://github.com/usuyama/ePillID-benchmark}}

\end{abstract}

\section{Introduction}

At least 1.5 million preventable adverse drug events (ADE) occur each year in the U.S.~\cite{IoM2006}. 
Medication errors related to ADEs can occur while writing or filling prescriptions, or even when taking or managing medications~\cite{aronson2009medication}. For example, pharmacists must verify thousands of pills dispensed daily in the pharmacy, a process that is largely manual and prone to error. Chronically-ill and elderly patients often separate pills from their original prescription bottles, which can lead to confusion and misadministration. %
Many errors are preventable; however, pills with visually similar characteristics are difficult to identify or distinguish, increasing error potential.%

\begin{figure}[t!]
    \begin{center}
    \includegraphics[width=0.8\linewidth]{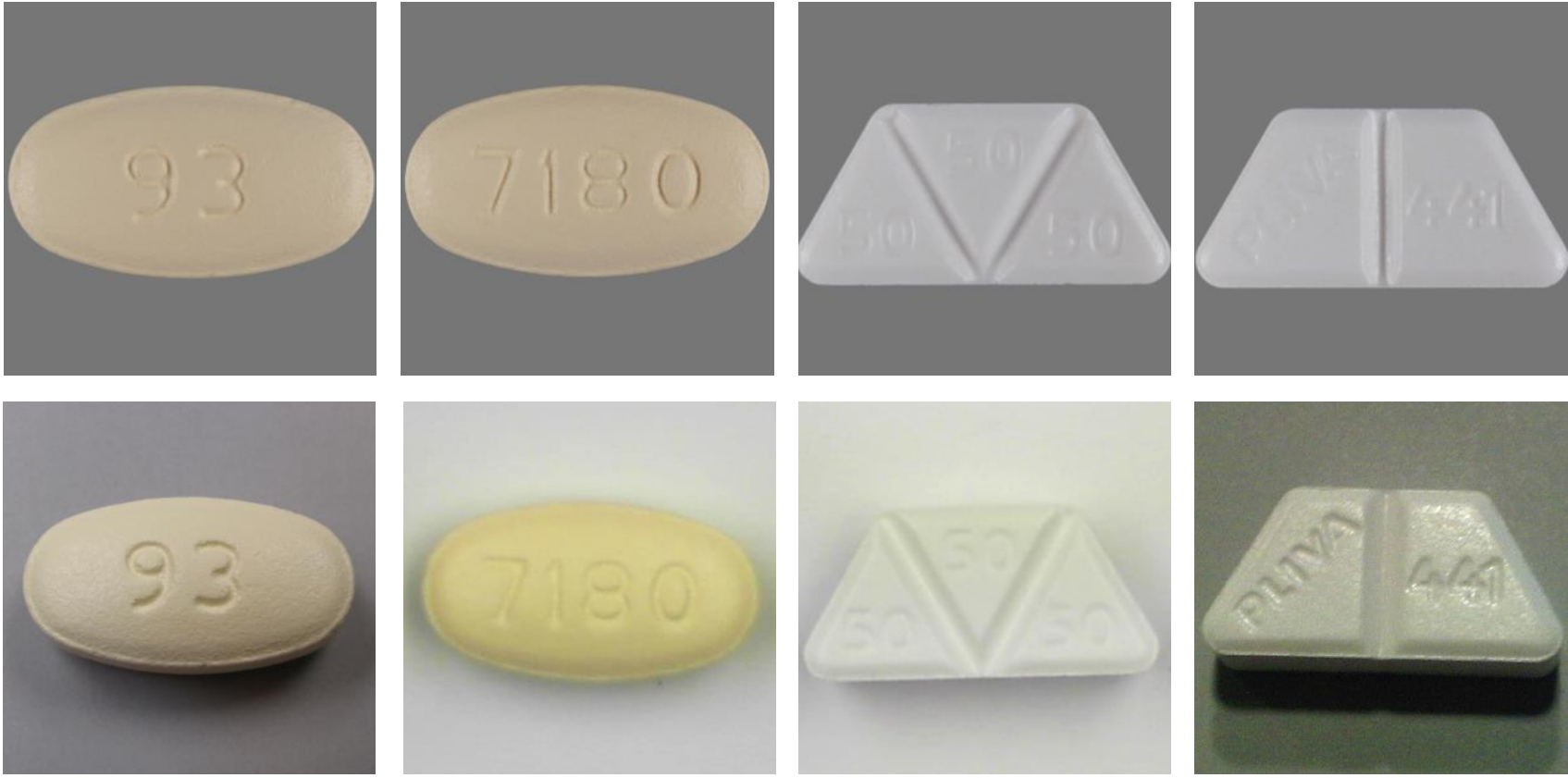}
    \end{center}
      \caption{\textbf{Representative examples of reference images (top row) and corresponding consumer images (bottom row)} Both sides (front and back) of the two pill types (NDC:00093-7180 and NDC:50111-0441, respectively) are shown. 
      }
      \label{fig:ref_cons_examples}
\end{figure}

The pill images available for vision-based approaches fall into two categories: reference and consumer images (Figure~\ref{fig:ref_cons_examples}). The reference images are taken with controlled lighting and backgrounds, and with professional equipment. For most of the pills, one image per side (two images per pill type) is available from the National Institutes of Health (NIH) Pillbox dataset~\cite{pillbox}. The consumer images are taken with real-world settings including different lighting, backgrounds, and equipment. Building pill-image datasets, especially for prescription medications, is costly and has additional regulatory obstacles as prescription medications require a clinician's order to be dispensed. A labeled dataset is scarce and even unlabeled images are hard to collect. This setup requires the model to learn representations from very few training examples. Existing benchmarks are either not publicly available or smaller scale, which are not suitable for developing real-world pill identification systems.

\begin{figure}[!t]
        \centering
        \includegraphics[width=0.8\linewidth]{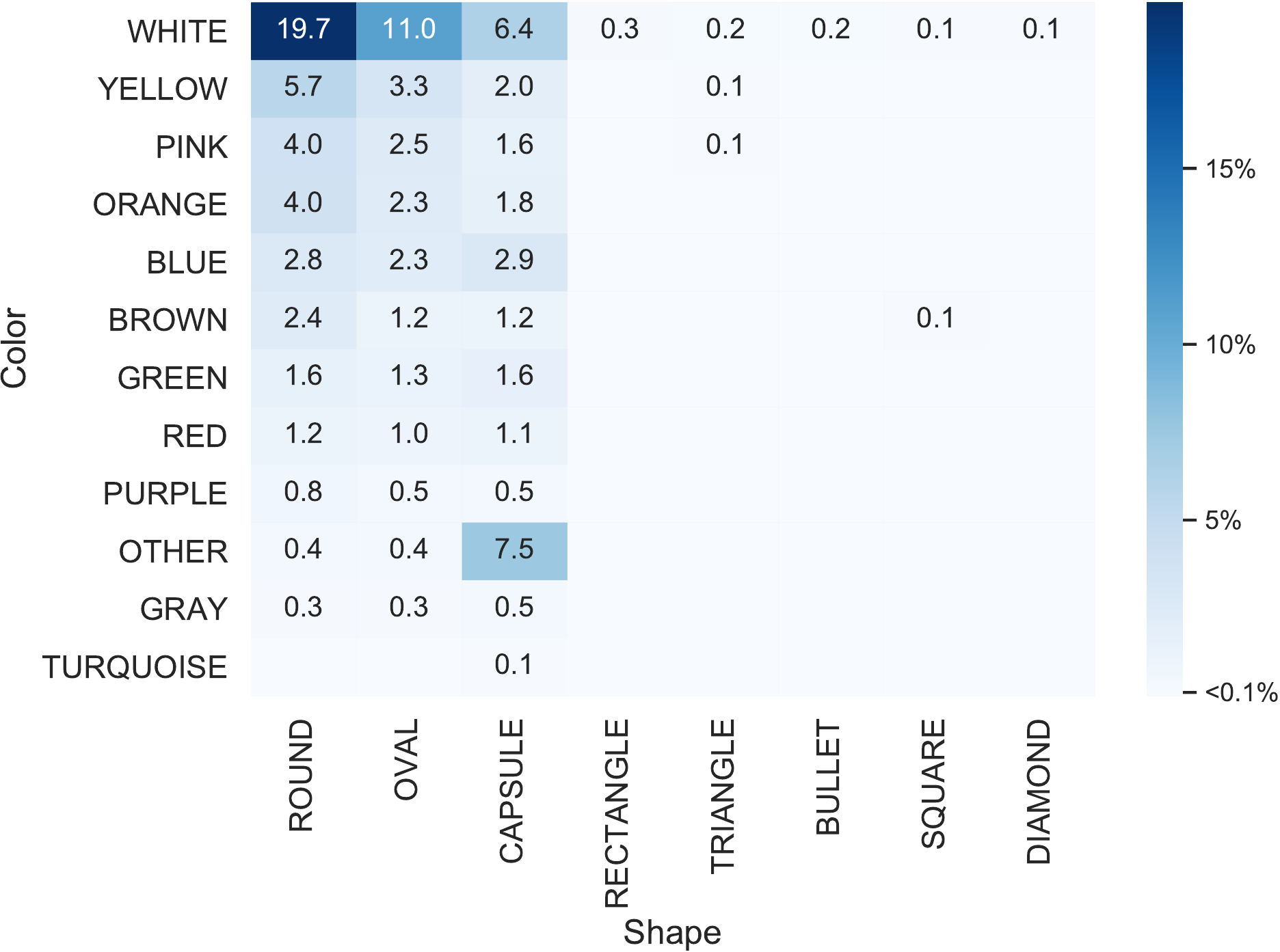}
        \caption{\textbf{The distribution of shape and color of pills.} For pills registered with multiple colors, we grouped them as OTHER. The combinations less than $0.1\%$ are omitted.}
        \label{fig:shape_color_heatmap}
\end{figure}

Over the years, deep learning has achieved unprecedented performance on image recognition tasks with efforts from both large-scale labeled datasets~\cite{imagenet,zhou2014learning} and modeling improvements~\cite{ResNet50CVPR2015,krizhevsky2012imagenet,simonyan2014very}; however, fine-grained visual categorization (FGVC) is still a challenging task, which needs to distinguish subtle differences within visually similar categories. FGVC tasks mainly include natural categories (e.g., birds~\cite{van2015building,wah2011caltech}, dogs~\cite{khosla2011novel} and plants~\cite{nilsback2008automated,wegner2016cataloging}) and man-made objects (e.g., cars~\cite{krause20133d,yang2015large} and airplanes~\cite{maji2013fine}). In this work, we target the pill identification task, which is an under-explored FGVC task yet important and frequent in healthcare settings. The distributions of pill shape and color are skewed to certain categories (Figure~\ref{fig:shape_color_heatmap}), highlighting the importance of distinguishing subtle differences such as materials, imprinted text and symbols.

The main contribution of this paper is introducing ePillID, a new pill identification benchmark with a real-world low-shot recognition setting. Leveraging two existing NIH datasets, our benchmark is composed of 13k images representing 9804 appearance classes (two sides for 4902 pill types). This is a low-shot fine-grained challenge because (1) for most of the appearance classes there exist only one image and (2) many pills have extremely similar appearances. %
Furthermore, we empirically evaluate various approaches with the benchmark to serve as baselines. 
The baseline models include standard image classification approaches and metric learning based approaches.
Finally, we present error analysis to motivate future research directions.

\section{Related Work}
\paragraph{Image-based Pill Identification:}

 In 2016, the NIH held a pill image recognition challenge~\cite{yaniv2016national}, and released a dataset composed of 1k pill types. The challenge winner~\cite{zeng2017mobiledeeppill} proposed a deep similarity-based approach~\cite{DeepRankingCVPR2014} and developed a mobile-ready version using knowledge-distillation techniques~\cite{hinton2015distilling}. Following the competition, classification-based approaches were applied, with higher recognition performance reported~\cite{delgado2019fast, szegedy2015going, wong2017development}. Aside from deep neural networks, various approaches with feature engineering have been proposed~\cite{caban2012automatic,cunha2014helpmepills,lee2012pill,neto2018cofordes,wong2017development}. For instance, the Hu moment~\cite{hu1962visual} was applied~\cite{cunha2014helpmepills,lee2012pill}, because of its rotation-invariant properties. Other methods~\cite{chen2013new,suntronsuk2016pill,yu2014pill,yu2015accurate} were proposed to generate imprint features, recognizing the importance of imprints for pill identification. The past methods achieved remarkable success; however, the lack of benchmarks prevents us from developing methods for real-world low-shot settings. Wong \textit{et al.}~\cite{wong2017development} created a 5284-image dataset of 400 pill types with a QR-like board to rectify geometric and color distortions. Yu \textit{et al.}~\cite{yu2014pill} collected 12,500 images of 2500 pill categories. Unfortunately, neither of these datasets are publicly available.

\paragraph{Fine-Grained Visual Categorization (FGVC):}

In many FGVC datasets~\cite{bossard2014food,beery2019iwildcam,kaur2019foodx,cui2018large}, the number of categories is not extremely large, often less than 200. Recent large-scale datasets ~\cite{imatproduct,fieldguide,van2018inaturalist} offer large numbers of categories with many images (e.g., \num{675170} training images for 5089 categories~\cite{van2018inaturalist}) with challenging long-tailed distributions. Compared to other FGVC benchmarks, the data distribution on the ePillID benchmark imposes a low-shot setting (one image for most of the classes) with a large number of classes (8k appearance classes). Among many algorithms~\cite{yang2018learning,wang2018learning,zheng2019looking} proposed for FGVC tasks, bilinear models~\cite{kong2017low,li2018towards,zheng2019learning} achieved remarkable performances by capturing higher-order interactions between feature channels. B-CNN~\cite{lin2015bilinear} is one of the first approaches, which obtains full bilinear features by calculating outer product at each location of the feature map, followed by a pooling across all locations; however, the full bilinear features can be very high dimensional (e.g., over 250k when the input has 512 channels). Compact Bilinear Pooling (CBP)~\cite{gao2016compact} addresses the dimensionality issue by approximating bilinear features with only a few thousand dimensions, which was shown to outperform B-CNN in few-shot scenarios.
Another line of work is metric learning~\cite{qian2015fine,cui2016fine,sun2018multi}, where an embedding space that captures semantic similarities among classes and images is learned. Metric learning has been also successfully used in few- and low-shot settings~\cite{wang2019cvpr,sungflood2018cvpr}, making it suitable for our ePillID benchmark. %

 \begin{figure}[t!]
    \begin{center}
    \includegraphics[width=0.8\linewidth]{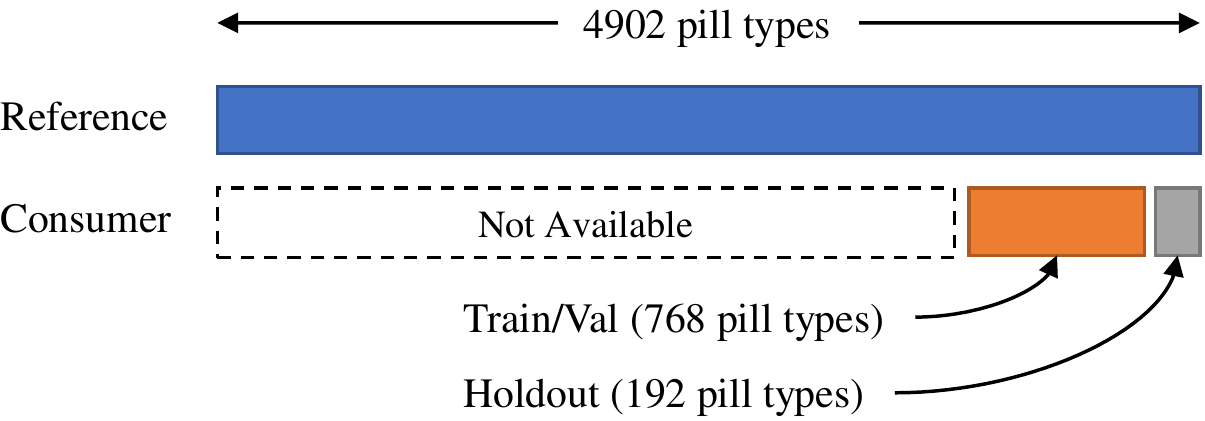}
    \end{center}
       \caption{\textbf{The distribution of reference and consumer images on the ePillID benchmark.} The reference images are available for 4902 pill types, while the consumer images are only available for 960 pill types. The dataset includes 9804 reference images (two images, front and back, for each pill type) and 3728 consumer images.}
    \label{fig:ePillID}
\end{figure}

\section{ePillID Benchmark}

We construct a new pill identification benchmark, ePillID, by leveraging the NIH challenge dataset~\cite{NIHpillcomp} and the NIH Pillbox dataset~\cite{pillbox}. We use the challenge dataset, which offers consumer images, as a base dataset and extend it with the reference images from the Pillbox dataset. In total, the ePillID dataset includes 3728 consumer images for 1920 appearance classes (two sides for 960 pill types) and 9804 reference images (two sides for 4902 pill types). This requires a fine-grained low-shot setup, where models have access to one reference image for all the 9804 appearance classes; however, there only exists a few consumer images for 1920 appearance classes (Figure~\ref{fig:ePillID}).

\paragraph{Experimental Setup:}

 The consumer images are split into 80\% training and 20\% holdout sets in such a way that the pill types are mutually exclusive. The training set is further split on pill types for 4-fold cross-validation. The models have access to reference images for all of the 4902 pill types, but consumer images are unavailable for most pill types during training. To evaluate the performance in situations where both front- and back-sides are available as input, we construct two-sided queries by enumerating all possible consumer image pairs for each pill type.

\paragraph{Evaluation Metrics:}

For each query, a model calculates an ordered list of pill types with confidence scores. 
Note that, for the experiments with both sides of a pill, a query consists of a pair of front and back images.
We consider Mean Average Precision (MAP) and Global Average Precision (GAP) for the model performance evaluation. %
For MAP, the average precision score is calculated separately for each query, and the mean value is calculated. MAP measures the ability to predict the correct pill types given queries. For GAP, all the query and pill-type pairs are treated independently, and the average precision score is calculated globally. GAP measures both the ranking performance and the consistency of the confidence scores i.e.\ the ability to use a common threshold across different queries~\cite{perronnin2009family}. We also calculate MAP@1 and GAP@1, where only the top pill type per query is considered. %

\begin{table}[!t]
    \centering
    \scalebox{0.65}{
\begin{tabular}{lcccc}
    \toprule
       Model &               GAP &             GAP@1 &               MAP &             MAP@1 \\
    \toprule
       \multicolumn{5}{c}{\bf{Plain Classification: Both-sides input}} \\
       \midrule
DenseNet121 &  $30.05 \pm 2.04$ &  $38.53 \pm 4.42$ &  $60.03 \pm 1.05$ &  $41.21 \pm 1.89$ \\
DenseNet201 &  $33.30 \pm 1.27$ &  $42.64 \pm 3.40$ &  $64.38 \pm 0.85$ &  $46.61 \pm 1.11$ \\
DenseNet161 &  $34.78 \pm 2.80$ &  $44.47 \pm 4.26$ &  $64.87 \pm 2.27$ &  $46.85 \pm 2.72$ \\
 DenseNet161 B-CNN &   $36.35 \pm 1.67$ &   $47.08 \pm 2.16$ &   $67.07 \pm 1.36$ &   $50.84 \pm 2.47$ \\
 DenseNet161 BCP &   $39.47 \pm 1.90$ &   $48.72 \pm 1.92$ &   $69.77 \pm 1.01$ &   $52.72 \pm 1.42$ \\
      \midrule          
         ResNet18 &  $24.13 \pm 1.99$ &  $33.13 \pm 2.34$ &  $54.70 \pm 2.13$ &  $35.70 \pm 2.79$ \\
         ResNet34 &  $29.07 \pm 2.45$ &  $38.53 \pm 2.54$ &  $58.88 \pm 2.26$ &  $40.34 \pm 2.01$ \\
         ResNet50 &  $35.16 \pm 3.46$ &  $45.43 \pm 4.22$ &  $65.79 \pm 1.85$ &  $48.32 \pm 2.34$ \\
         ResNet50 B-CNN &  $38.15 \pm 2.73$ &  $48.13 \pm 4.17$ &  $68.68 \pm 2.23$ &  $51.92 \pm 3.23$ \\
         ResNet50 CBP &  $38.97 \pm 2.74$ &  $48.63 \pm 3.94$ &  $68.35 \pm 3.08$ &  $51.60 \pm 4.24$ \\
         ResNet101 &  $36.15 \pm 3.21$ &  $46.96 \pm 5.29$ &  $65.50 \pm 2.95$ &  $48.20 \pm 5.02$ \\
         ResNet152 &  $39.57 \pm 1.23$ &  $49.97 \pm 1.58$ &  $68.51 \pm 1.09$ &  $51.64 \pm 2.31$ \\
         ResNet152 B-CNN &  $44.75 \pm 2.11$ &  $55.14 \pm 3.67$ &  $72.65 \pm 1.69$ &  $56.75 \pm 2.51$ \\
         ResNet152 CBP &  $ 41.45 \pm 2.15$ &     $52.43 \pm 2.07$ &     $69.12 \pm 2.10$ &     $52.64 \pm 2.29$ \\
     \midrule
    \multicolumn{5}{c}{\bf{Multi-head Metric Learning: Both-sides input}} \\
     \midrule
        DenseNet121 &  $74.04 \pm 2.32$ &  $90.15 \pm 2.63$ &  $93.42 \pm 0.42$ &  $87.58 \pm 0.87$ \\
        DenseNet201 &  $74.19 \pm 2.08$ &  $87.69 \pm 2.58$ &  $92.52 \pm 0.91$ &  $86.42 \pm 1.56$ \\
        DenseNet161 &  $76.60 \pm 2.35$ &  $88.86 \pm 2.88$ &  $93.41 \pm 0.72$ &  $87.78 \pm 1.55$ \\
        DenseNet161 B-CNN &  $75.27 \pm 4.65$ &  $88.24 \pm 2.83$ &  $93.11 \pm 0.81$ &  $87.22 \pm 1.57$ \\
        DenseNet161 BCP &  $77.15 \pm 2.89$ &  $89.11 \pm 2.36$ &  $94.03 \pm 1.03$ &  $88.98 \pm 1.78$ \\
        \midrule
         ResNet18 &  $66.58 \pm 1.91$ &  $87.87 \pm 2.86$ &  $91.16 \pm 0.75$ &  $84.38 \pm 1.36$ \\
         ResNet34 &  $73.61 \pm 3.17$ &  $89.22 \pm 3.35$ &  $92.74 \pm 1.95$ &  $87.10 \pm 2.96$ \\
         ResNet50 &  $78.27 \pm 2.35$ &  $90.71 \pm 2.55$ &  $94.02 \pm 0.66$ &  $89.06 \pm 1.05$ \\
         ResNet50 B-CNN &  $79.95 \pm 1.96$ &  $90.26 \pm 1.56$ &  $94.59 \pm 0.23$ &  $89.66 \pm 0.62$ \\
         ResNet50 CBP &  $78.44 \pm 1.48$ &  $91.36 \pm 2.83$ &  $95.27 \pm 0.43$ &  $91.01 \pm 0.97$ \\
         ResNet101 &  $79.83 \pm 2.11$ &  $92.23 \pm 1.12$ &  $94.99 \pm 0.40$ &  $90.65 \pm 0.60$ \\
         ResNet152 &  $78.63 \pm 2.75$ &  $90.54 \pm 1.96$ &  $95.71 \pm 0.83$ &  $91.93 \pm 1.56$ \\
         ResNet152 B-CNN &  $80.45 \pm 1.09$ &  $89.61 \pm 1.83$ &  $95.01 \pm 0.43$ &  $90.65 \pm 0.81$ \\
         ResNet152 CBP &  $81.20 \pm 1.47$ &  $91.19 \pm 0.28$ &  $95.76 \pm 0.40$ &  $92.01 \pm 0.67$ \\
       \midrule
       \multicolumn{5}{c}{\bf{Multi-head Metric Learning: Single-side input}} \\
       \midrule
        ResNet34 &  $54.70 \pm 2.58$ &  $78.69 \pm 2.37$ &  $80.52 \pm 1.76$ &  $70.61 \pm 2.11$ \\
          ResNet50 &  $61.75 \pm 1.40$ &  $81.63 \pm 2.15$ &  $82.34 \pm 0.79$ &  $72.56 \pm 1.35$ \\
         ResNet101 &  $63.93 \pm 1.71$ &  $84.51 \pm 1.68$ &  $84.18 \pm 1.23$ &  $75.72 \pm 1.48$ \\
         ResNet152 &  $64.61 \pm 2.45$ &  $82.67 \pm 1.17$ &  $85.25 \pm 0.91$ &  $76.42 \pm 1.54$ \\
       \bottomrule
    \end{tabular}
    }
    \caption{\textbf{Recognition results on the ePillID benchmark.} Mean and standard deviations of the holdout metrics from the 4-fold cross-validation are reported in percentages. }
    \label{tab:results}
\end{table}

\section{Experiments}
\label{sec:experiments}
We first introduce our baseline approaches, then present quantitative and qualitative results.

\subsection{Baseline Models}

We use ResNet~\cite{ResNet50CVPR2015} and DenseNet~\cite{densenet} as base networks pretrained on ImageNet~\cite{imagenet} for initial weights. In addition to the global average pooling layer as features, we evaluate two bilinear methods, B-CNN~\cite{lin2015bilinear} and CBP~\cite{gao2016compact}, applied to the final pooling layer of the base network. For CBP, we use their Tensor Sketch projection with 8192 dimensions, which was suggested for reaching
close-to maximum accuracy. We insert a 1x1 convolutional layer before the pooling layer to reduce the dimensionality to 256.

\paragraph{Plain Classification:}

As a first set of baselines, we train the models with the standard softmax cross-entropy loss. For regularization, we add a dropout layer~\cite{srivastava2014dropout} with probability 0.5 before the final classification layer. We use the appearance classes as target labels during training, and take the max of the softmax scores for calculating pill-type confidence scores. For the two-sides evaluation, the mean confidence score is used for a score between a two-sided input and a pill type. 

\paragraph{Multi-head Metric Learning:}\label{sec:metric_losses}

As another set of baselines, we employ a combination of four losses to learn an embedding space optimized for fine-grained low-shot recognition.
The set up is a multi-task training procedure:
\begin{equation}
L_{final} = \lambda_{SCE} L_{SCE} + \lambda_{\eta} L_{\eta} + \lambda_{\rho} L_{\rho} + \lambda_{\Gamma} L_{\Gamma},
\end{equation}
where $L_{SCE}$ indicates softmax cross-entropy, $L_{\eta}$ cosine-softmax loss (ArcFace~\cite{deng2019cvpr}), $L_{\rho}$ triplet loss~\cite{schroff2015cvpr}, and $L_{\Gamma}$ contrastive loss~\cite{chopra2005learning}. 
The loss weights, $\lambda_{SCE}$, $\lambda_{\eta}$, $\lambda_{\rho}$, and $\lambda_{\Gamma}$
are chosen empirically with a ResNet50 model~(Section~\ref{sec:impl}).
In order to compute the loss for every mini-batch, the triplet and contrastive loss requires additional sampling and pairing procedures. We apply online hard-example mining to find informative negatives for the triplets and pairs respectively evaluated by these losses. %
The trained model is used for generating embeddings for the query consumer images and all the reference images. We calculate cosine similarities between the query and reference embeddings and use them as confidence scores.

\paragraph{Implementation Details:}\label{sec:impl}

The images are cropped and resized to $224 \times 224$. Extensive data augmentation is employed to mimic consumer-image-like variations, including rotation and perspective transformation. The Adam optimizer~\cite{adam2014} is used with an initial learning rate of $\num{1e-4}$. The learning rate is halved whenever a plateau in the validation GAP score is detected. The model hyper-parameters are chosen to optimize the average validation GAP score using the 4-fold cross-validation. The loss weights for the metric learning are determined based on the ResNet50 experiment~(Table~\ref{tab:loss_weights}). The training is done with the mini-batch of 48 images in a machine equipped with Intel Xeon E5-2690, 112GB RAM, one NVIDIA Tesla P40, CUDA 8.0 and PyTorch 0.4.1. 

\begin{table}[t!]
    \centering
    \scalebox{0.65}{
\begin{tabular}{ccccc}
\toprule
Contrastive~($\lambda_{\Gamma}$) &  Triplet~($\lambda_{\rho}$) &  SCE~($\lambda_{SCE}$) &  CSCE~($\lambda_{\eta}$) &    Validation GAP \\
\midrule
             1.0 &          1.0 &        1.0 &     0.1 &  $77.18 \pm 1.90$ \\
             1.0 &          1.0 &        1.0 &     0.3 &  $74.07 \pm 2.29$ \\
             1.0 &          1.0 &        1.0 &     0.0 &  $75.48 \pm 1.02$ \\
             1.0 &          0.0 &        1.0 &     0.0 &  $74.40 \pm 2.93$ \\
             0.0 &          1.0 &        1.0 &     0.0 &  $73.89 \pm 2.16$ \\
             1.0 &          1.0 &        0.0 &     0.1 &  $71.38 \pm 4.52$ \\
             0.0 &          0.0 &        1.0 &     0.0 &  $61.78 \pm 3.24$ \\
\bottomrule
\end{tabular}
}
\caption{\textbf{Validation GAP scores from the corss-validation for comparing the metric-learning loss weights.} 
The ResNet50 baseline model is used.}
\label{tab:loss_weights}
\end{table}

\subsection{Quantitative Results}

In Table~\ref{tab:results}, we report the baseline results on the ePillID benchmark. The plain classification baselines performed much worse than the metric learning baselines, suggesting the difficulty of the low-shot fine-grained setting. The performance differences among the models are consistent with the ImageNet pretraining performance~\cite{bianco2018benchmark}. The multi-head metric leraning models performed remarkably well, achieving over 95\% MAP and 90\% GAP@1. In most of the cases, bilinear pooling methods outperformed the global average pooling counterpart, showing the representation power of the bilinear features. As an ablation study, we report single-side experiment results i.e.\ only one image per query. The ResNet152 metric learning approach achieved over 85\% MAP and 82\% GAP@1; however, the results indicate that both sides are required for accurate identification.

\subsection{Qualitative Results}

\begin{figure}[!t]
	\begin{center}
		\includegraphics[width=0.9\linewidth]{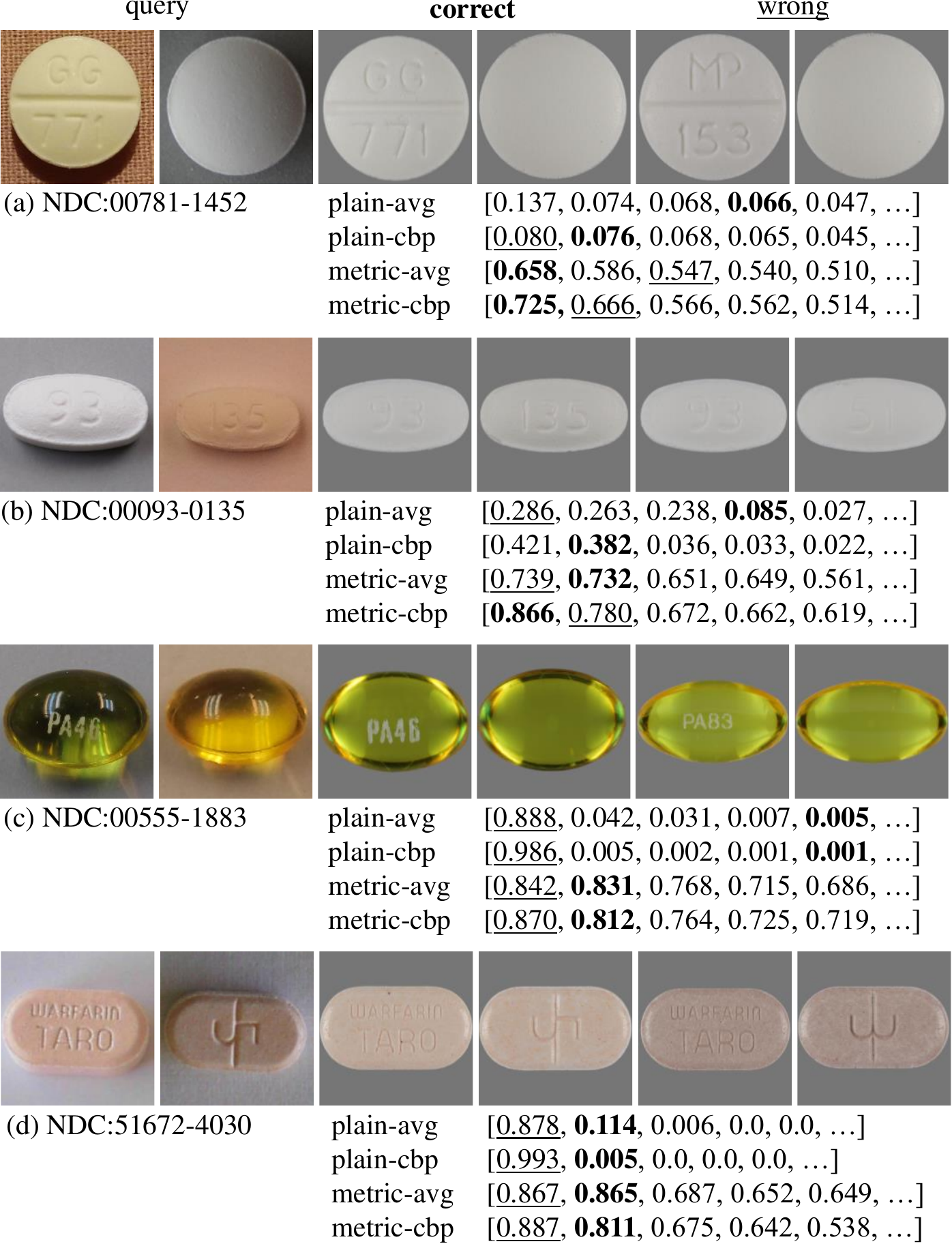}
	\end{center}
	\caption{\textbf{Qualitative results from the ePillID holdout dataset.} For each query, top confidence scores are shown from the four ResNet152 baselines. The reference images of \textbf{correct} and \underline{wrong} pill types are shown with the input consumer images in the left.}
	\label{fig:qualitative}
\end{figure}

Figure~\ref{fig:qualitative} depicts qualitative comparisons from examples of the ePillID holdout dataset with confidence scores. In (a), the plain classification approaches misclassified, whereas the metric learning approaches identified successfully, even with the challenging lighting and background variations in the query images. In (b), only the metric learning with CBP approach identified correctly, suggesting CBP was effective for capturing the small difference in the imprinted text. For (c) and (d), all the four models failed to identify the correct types. In (c), the consumer images are affected by the lighting variations with the shiny pill material. In (d), the pill types share extremely similar appearances, except the one character difference in the imprinted text.

\section{Conclusion}

We introduced ePillID, a low-shot fine-grained benchmark on pill image recognition. To our knowledge, this is the first publicly available benchmark that can be used to develop pill identification systems in a real-world low-shot setting. We empirically evaluated various baseline models with the benchmark. The multi-head metric learning approach performed remarkably well; however, our error analysis suggests that these models still cannot distinguish confusing pill types reliably. 
In the future, we plan to integrate optical character recognition~(OCR) models. OCR integration has been explored for storefronts and product FGVC tasks~\cite{karaoglu2016words,bai2018integrating} and recent advances in scene text recognition are promising~\cite{karatzas2015icdar,gomez2017icdar2017,liu2018fots}; however, existing OCR models are unlikely to perform reliably on pills as they stand. Challenging differences include low-contrast imprinted text, irregular-shaped layouts, lexicon, and pill materials such as capsules and gels. 
Finally, we plan to extend the benchmark further with more pill types and images as we collect more data. By releasing this benchmark, we hope to support further research in this under-explored yet important task in healthcare.

{\small
\bibliographystyle{ieee_fullname}
\bibliography{pillbib}
}

\end{document}